\documentclass[letterpaper, 10 pt, conference]{ieeeconf}  

\IEEEoverridecommandlockouts                              

\usepackage{amsmath}
\usepackage{xcolor, soul}
\usepackage{subcaption}
\usepackage{amsmath}
\usepackage{amsfonts}
\usepackage{amssymb}
\usepackage{bm}
\usepackage{multicol}
\usepackage{multirow}
\usepackage{optidef}
\usepackage{scalerel}
\usepackage{tikz}
\usetikzlibrary{svg.path}
\usepackage{colortbl}
\usepackage{color}
\definecolor{celadon}{rgb}{0.67, 0.88, 0.69}
\usepackage{algorithm}
\usepackage{algpseudocode}
\usepackage{graphicx}
\usepackage{textcomp}
\newcommand\BibTeX{{\rmfamily B\kern-.05em \textsc{i\kern-.025em b}\kern-.08em
T\kern-.1667em\lower.7ex\hbox{E}\kern-.125emX}}

\definecolor{orcidlogocol}{HTML}{A6CE39}
\tikzset{
  orcidlogo/.pic={
    \fill[orcidlogocol] svg{M256,128c0,70.7-57.3,128-128,128C57.3,256,0,198.7,0,128C0,57.3,57.3,0,128,0C198.7,0,256,57.3,256,128z};
    \fill[white] svg{M86.3,186.2H70.9V79.1h15.4v48.4V186.2z}
                 svg{M108.9,79.1h41.6c39.6,0,57,28.3,57,53.6c0,27.5-21.5,53.6-56.8,53.6h-41.8V79.1z M124.3,172.4h24.5c34.9,0,42.9-26.5,42.9-39.7c0-21.5-13.7-39.7-43.7-39.7h-23.7V172.4z}
                 svg{M88.7,56.8c0,5.5-4.5,10.1-10.1,10.1c-5.6,0-10.1-4.6-10.1-10.1c0-5.6,4.5-10.1,10.1-10.1C84.2,46.7,88.7,51.3,88.7,56.8z};
  }
}

\newcommand\orcidiconKLW[1]{\href{https://orcid.org/0000-0002-1938-4222}{\mbox{\scalerel*{
\begin{tikzpicture}[yscale=-1,transform shape]
\pic{orcidlogo};
\end{tikzpicture}
}{|}}}}

\newcommand\orcidiconAJP[1]{\href{https://orcid.org/0000-0002-8864-9044}{\mbox{\scalerel*{
\begin{tikzpicture}[yscale=-1,transform shape]
\pic{orcidlogo};
\end{tikzpicture}
}{|}}}}

\newcommand\orcidiconHYC[1]{\href{https://orcid.org/0000-0003-1074-4225}{\mbox{\scalerel*{
\begin{tikzpicture}[yscale=-1,transform shape]
\pic{orcidlogo};
\end{tikzpicture}
}{|}}}}

\usepackage{hyperref}

\title{\LARGE \bf
A Modular, Tendon Driven Variable Stiffness Manipulator with Internal Routing for Improved Stability and Increased Payload Capacity
}

\author{Kyle L. Walker$^{1*}$ \orcidiconKLW{0000-0002-1938-4222}, Alix J. Partridge$^{1*}$ \orcidiconAJP{0000-0002-8864-9044}, Hsing-Yu Chen$^{1*}$ \orcidiconHYC{0000-0003-1074-4225}, Rahul R. Ramachandran$^{1}$, Adam A. Stokes$^{2}$, \\ Kenjiro Tadakuma$^{3}$, Lucas Cruz da Silva$^{4}$ and Francesco Giorgio-Serchi$^{2}$
\thanks{This work was supported by Brazilian Agency for Industrial Research and Technological Innovation (EMBRAPII).}
\thanks{*Authors contributed equally to this research.}
\thanks{$^{1}$Kyle L. Walker$^{1}$, Alix J. Partridge, Hsing-Yu Chen and Rahul Ramachandran are with the National Robotarium,
        Boundary Road North, Heriot Watt University, Edinburgh, U.K.
        }%
\thanks{$^{2}$Adam A. Stokes and Francesco Giorgio-Serchi are with the Institute for Integrated Micro and Nano Systems, University of Edinburgh,
        Edinburgh, U.K. Correspondence: {\tt\small f.giorgio-serchi@ed.ac.uk}}
\thanks{$^{3}$Kenjiro Tadakuma is with the Tough Cyberphysical AI Research Center (TCPAI), Tohoku University,
        Sendai, Japan.}%
\thanks{$^{4}$Lucas Cruz da Silva is with the Robotics Department, Senai Cimatec, Salvador, Bahia, Brazil.}%
}

\begin{document}

\maketitle
\thispagestyle{empty}
\pagestyle{empty}

\begin{abstract}

Stability and reliable operation under a spectrum of environmental conditions is still an open challenge for soft and continuum style manipulators. The inability to carry sufficient load and effectively reject external disturbances are two drawbacks which limit the scale of continuum designs, preventing widespread adoption of this technology. To tackle these problems, this work details the design and experimental testing of a modular, tendon driven bead-style continuum manipulator with tunable stiffness. By embedding the ability to independently control the stiffness of distinct sections of the structure, the manipulator can regulate it's posture under greater loads of up to 1kg at the end-effector, with reference to the flexible state. Likewise, an internal routing scheme vastly improves the stability of the proximal segment when operating the distal segment, reducing deviations by at least 70.11\%. Operation is validated when gravity is both tangential and perpendicular to the manipulator backbone, a feature uncommon in previous designs. The findings presented in this work are key to the development of larger scale continuum designs, demonstrating that flexibility and tip stability under loading can co-exist without compromise.

\end{abstract}

\section{INTRODUCTION}

The manipulation of objects is a process that is required across all spheres of the earth, hydro- \cite{Galloway2016SoftReefs}, bio- \cite{Gombert_2014}, atmos- and beyond \cite{hiltz2001canadarm}, with each sphere presenting its own environmental constraints and challenges that must be overcome. One of the most pressing environmental factors is that of pressure, which transitions from vacuum in the outer atmosphere, to almost 110 MPa at the deepest point in the ocean \cite{Li_Chen_2021}, severely limiting what forms of actuation can be utilized. As such, systems that require internal chambers, such as those that are pneumatically driven \cite{Robertson2017NewCapabilities, Faudzi2018SoftActuator} are typically avoided in favour of tendon driven systems \cite{Stuart2017TheManipulation, Fang2023DesignSystem} or fully motorized designs \cite{Rollinson2014DesignRobot}. 

\begin{figure}[t!]
    \centering
    \includegraphics[width=\linewidth]{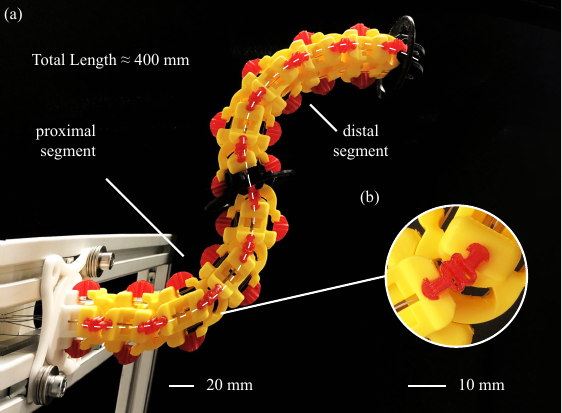}
    \caption{The modular manipulator design (a) in a horizontal configuration, highlighting the flexible hinges (b). Black beads indicate the tip of each segment and have increased radial surface area for the addition of motion tracking markers to aid motion capture. }
    \label{application_fig}\vspace{-0.5cm}
\end{figure}

\begin{figure*}[t!]
    \centering
    \includegraphics[width=\linewidth]{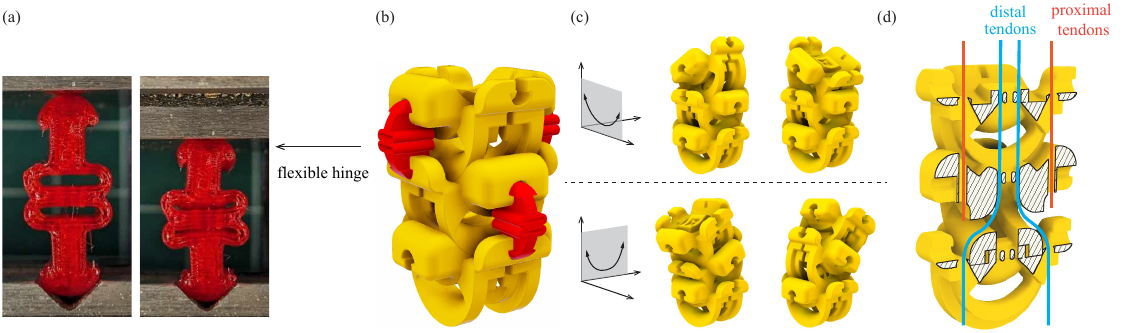}
    \caption{Overview of the modular design, showing the (a) compression of flexible hinges, (b) construction of successive beads, (c) discretised motion of successive beads and (d) internal routing of the distal tendons at the transition point between segments.}
    \label{design}\vspace{-0.5cm}
\end{figure*}

Manipulators that operate outside of controlled environments, such as warehouses or factories, should also be robust to the dynamics of the outside world. Examples from the field of soft robotics, and similarly compliant continuum devices \cite{Chen2022AOpportunities} pose a potential solution to this robustness, as they are composed of materials and structures that are able to resist or conform to change, without the need for high level control \cite{Pfeifer2009MorphologicalEnvironment}. Though structurally advantageous, compliant systems are typically more difficult to control in a repeatable and stable manner \cite{GeorgeThuruthel2018ControlSurvey} and, most importantly, they suffer from reduced load bearing capabilities when compared with their more traditional robotic counterparts \cite{Kumar_Singh_Krishna_2014}. This crucially constrains the manipulation ability during highly dynamic tasks or when subject to external disturbances,  as well as drastically limiting the size of soft manipulators. Size constraints and poor load-bearing capability are two significant factors that prevent widespread adoption of soft manipulators in real-world applications \cite{Chen2022AOpportunities}. 


Recent research into the use of localised stiffening has shown success for increased payload capacity in soft robots. It also shows promise as a method for improving the stability of soft systems \cite{Bruder2023IncreasingStiffening}; however, the proposed system utilises pneumatic actuation, which is unsuitable for manipulation across mediums with broadly differing nominal hydrostatic pressure. Other methods for stiffening soft and compliant devices include jamming beads to lock the manipulator into position through friction \cite{Tadakuma2020Fire-ResistantMechanism} and layer jamming \cite{Mukaide2020Radial-LayerConfiguration}, but these have not been developed for internal actuation, requiring an external intervention in acquiring the desired configuration. Cable driven stiffening has also been presented that adds additional cables helically to a manipulator that can then be commanded to drive an increased stiffness without effecting the actuating cables \cite{Molaei2023IndependentCoupling}. Such a system has only been proven in a planar manipulator, but similar concepts could be used for spatial manipulators.
Observation of biological systems has also led to the creation of bio-inspired components and structures that improve the capabilities of tendon driven manipulators \cite{Liu2023AnGrasping, Cianchetti2014BioinspiredAlloys}. 

With reference to the above, this work presents a multi-segment, tendon driven variable stiffness manipulator design, adopting a modular approach to facilitate retrofitting, repair and scalability, Fig. \ref{application_fig}. The stiffening element of the manipulator, which allows for independent stiffening of different sections, is designed to both improve stability at the proximal segment during highly dynamic operation of the distal segment, as well as improve load bearing capacity. In addition, an internal tendon routing scheme is adopted to further enhance the stability factor. The use of compliant hinges mimics flexible joints and assists in stiffness modulation. Stiffening performance is assessed by evaluating the deflection of the proximal segment during actuation of the distal segment across different tendon routing schemes. The stability of the tip under loading is tested with the manipulator operating tangential and perpendicular to gravity, validating the manipulators ability to withstand gravitational effects. The results support the claim that these traits can facilitate the development of longer continuum style manipulators, whilst retaining the benefits of the compliant design. 

\section{MANIPULATOR DESIGN}

The design of the manipulator is displayed in Fig. \ref{design}; each component shown in Fig. \ref{design}(a)-(d) is described in detail in the following sections.
%

%

\subsection{Discretised Bead Concept}

A modular design was selected, allowing the user to change the length depending on their task-specific requirements. As such, inspiration was taken from prior work that uses stacked beads to build the system to a given length while providing a mechanism for stiffening \cite{Tadakuma2020Fire-ResistantMechanism, Onda2023TubePressure}. 

Similar to
\cite{Fang2023DesignSystem, Lin2022DesignUsage}, a discrete backbone structure was chosen with rigid beads used as the constitutive units. The beads, depicted in Fig. \ref{design}(b), encompass a circular framed structure which ensured relative rotation between one bead and the following, Fig. \ref{design}(c). Relative sliding of successive beads is enabled  by a channel at each bead base that interlocks with the top of the neighbouring bead, Fig. \ref{design}(b). Each consecutive pair of beads is arranged with a 90$^\circ$ offset between them, such that the top of the bead and base of the bead are granted planar motion in orthogonal directions, Fig. \ref{design}(c), (d). This endows the assembled sequence of beads with spatial mobility, whilst retaining the benefits of discrete, planar interactions between each bead.  

The beads are fabricated from FDM printed PLA with 15\% infill to reduce weight. Each bead is 26mm long (measured from the hinge socket to the top of the circular head) and 38mm wide, weighing $\approx 7$g. In total, the manipulator presented here consists of 20 beads, amounting to a constructed length of $\approx$ 400mm in a flexible state, Fig. \ref{application_fig}.


\subsection{Flexible Hinges}

Flexible hinges with a sprung centre, Fig. \ref{design}(a), were specifically designed to couple successive beads together via two hinges, one on either side of the bead,  Fig. \ref{application_fig}(b) and Fig. \ref{design}(b). These elastic hinges further enable planar motion between neighbouring beads and ensure beads are separated. This separation prevents shearing between consecutive beads during actuation, removing friction and reducing the force required to operate the manipulator. When friction is required for stiffening the manipulator, all actuating cables are tightened and the sprung centres of the hinges compress, allowing the beads to contact and jam together, as shown in Figures \ref{application_fig}(b) and \ref{design}(a). This passively enables a bi-modal operation of the system, enabling the transition from a highly mobile state to a stiffened state when tendons are actuated concurrently.

The hinges are fabricated from FDM printed TPU (Shore hardness 95A) with 15\% infill and slot into holes within the beads, maintaining the modular nature of the manipulator. Each hinge is $\approx$19mm in height and weighs $<$1g.

\subsection{Internal Routing}

Typically, tendons are routed as far as possible from the backbone of the manipulator, either through holes with increasing radii for each segment or by positioning holes around the circumference of the manipulator \cite{Webster_Jones_2010}. To decouple tangential tension of the stiffened segments from the bending moment of the unstiffened segments, tendons were routed through the centre of the manipulator before transitioning to the outside for the segments they were manipulating, Fig. \ref{design}(d). This arrangement reduces the deflection of previous segments during manipulation, by limiting the distributed force applied by the actuating tendon. This idea is further explored in Section \ref{results} through physical experimentation and is further detailed in the literature \cite{Kim_2017}.

To achieve internal tendon routing, the beads were designed with a cut away section in the arching walls to allow the tendons to pass through at a gentle angle, thereby reducing stress and minimising friction, as presented in Figure \ref{application_fig} and \ref{design}(d). Tendons are then channelled into holes at the bead base to prevent them from interfering.    


\begin{figure}[t!]
    \centering
    \includegraphics[width=0.8\linewidth]{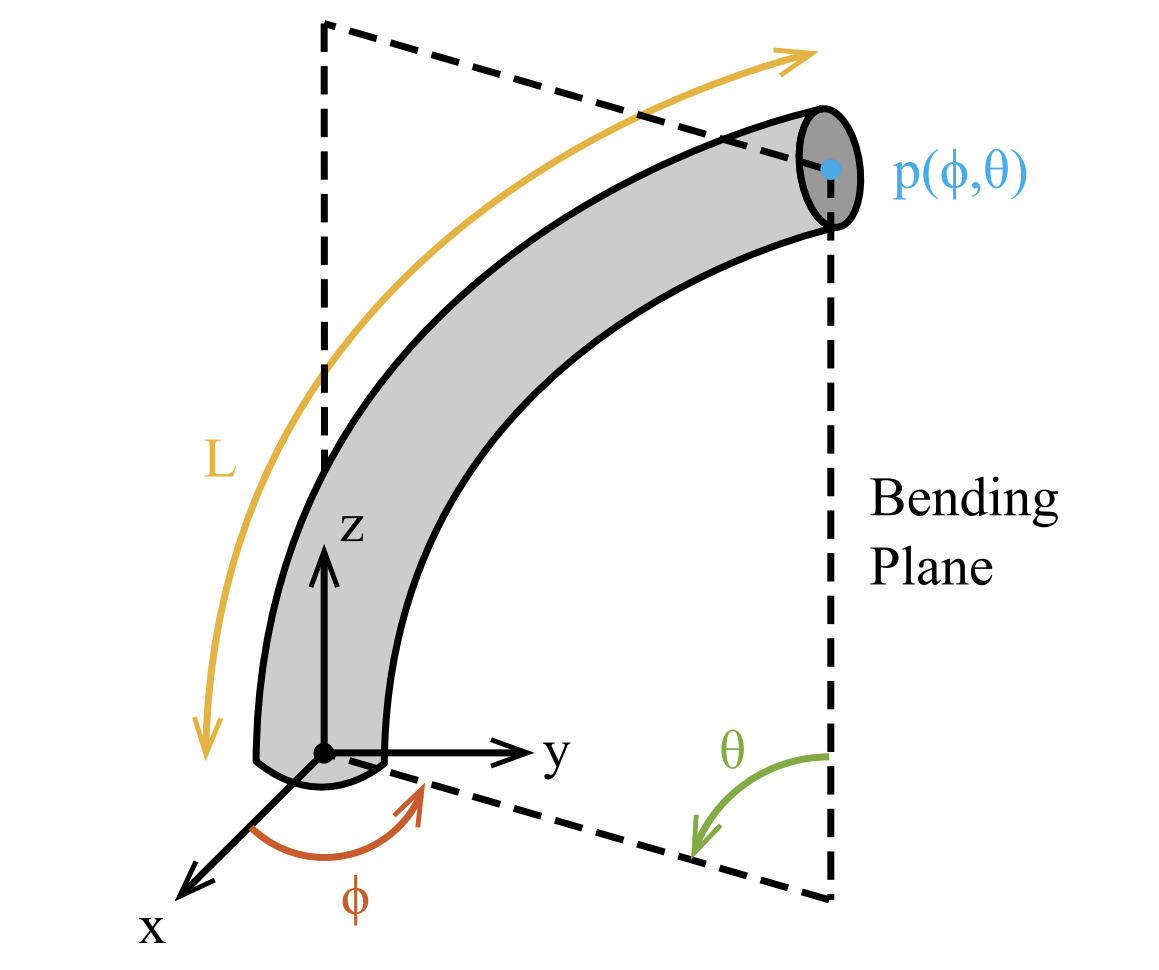}
    \caption{Piecewise Constant Curvature (PCC) kinematic model of a single spatial segment, where the dashed lines represent the bending plane. }
    \label{pcc_kinematics}\vspace{-0.5cm}
\end{figure}

\section{KINEMATIC MODELLING}

\begin{figure}[t!]
    \centering
    \includegraphics[width=\linewidth]{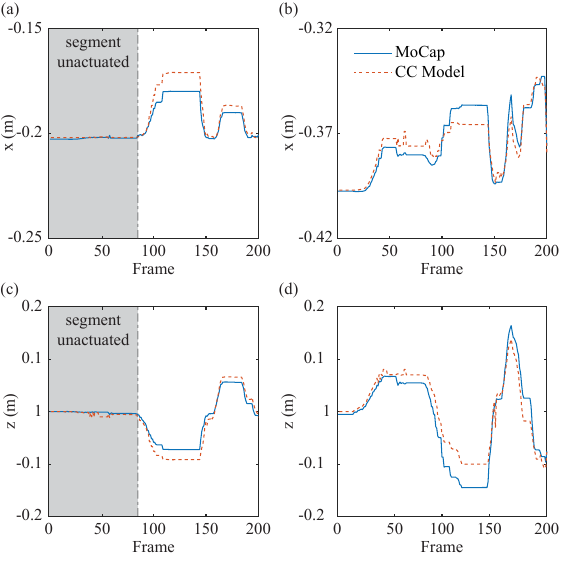}
    \caption{PCC behaviour validation of the manipulator design obtained via the MoCap. Shown is the recorded and modelled planar tip positions for the (a)(c) proximal segment and (b)(d) distal segment, relative to the base.}\vspace{-0.5cm}
    \label{pcc_proof}
\end{figure}

The manipulator consists of two independently actuated segments which can each be modelled according to a Piecewise Constant Curvature (PCC) approximation \cite{Webster_Jones_2010, Rucker_Webster_III_2011} (see Figure \ref{pcc_kinematics}). The actuator is assumed to undergo no twist and no elongation during operation; the change in length during stiffening is minimal ($<2\%)$ and twisting is resisted structurally by the interface between each bead and the hinges. As such the assumption of PCC behaviour is valid as is demonstrated in Fig. \ref{pcc_proof}. Therefore, the configuration variable for the $i$-th segment is formed of two angular variables, namely the bending plane angle $\phi_i$ and curvature angle $\theta_i$, such that $q_i = [\phi_i, \theta_i]^T \in \mathbb{R}^2$.

It follows that the tip position, $p_i \in \mathbb{R}^3$, of each segment can be described by the following kinematic relationship:
\begin{equation}
    p_i(\phi_i, \theta_i) =  \frac{L_i}{\theta_i} \left[ c\phi_i(1-c\theta_i) \quad s\phi_i(1-c\theta_i) \quad s\theta_i \right] ^T
\end{equation}
where $L_{i}$ is the segment length; $c(\cdot)$ and $s(\cdot)$ are shorthand for $\cos(\cdot)$ and $\sin(\cdot)$. Similarly, the following transformation matrix can be derived for propagation between reference frames with respect to the base frame of the segment (in this instance segment tip position is our reference frame):
\begin{equation}
    T_{i}(\phi_i, \theta_i) = \begin{bmatrix}
        R_z(\phi_i)R_y(\theta_i)R_z(-\phi_i) & p_i(\phi_i, \theta_i) \\
        0 & 1
    \end{bmatrix}
\end{equation}
where $R_z$ and $R_y$ refer to rotation's about the $z$ and $y$ axes. This facilitates deduction of the manipulator end-effector position through knowledge of the configuration variable, extracted using an OptiTrack  Motion Capturing System (MoCap) in this work.

\section{EXPERIMENTAL CHARACTERISATION} \label{results}

To validate operation of the design and the implementation of the aforementioned features, a physical manipulator was manufactured and tested experimentally (see Fig. \ref{application_fig}). Throughout the experiments, a 4 camera OptiTrack MoCap was used to analyse both tip rotations and positions of the proximal and distal segments. To realise stiffening through jamming, a tensile load of 30N was applied to each tendon through operation of the motors. 

\begin{figure}[t!]
    \centering
    \includegraphics[width=\linewidth]{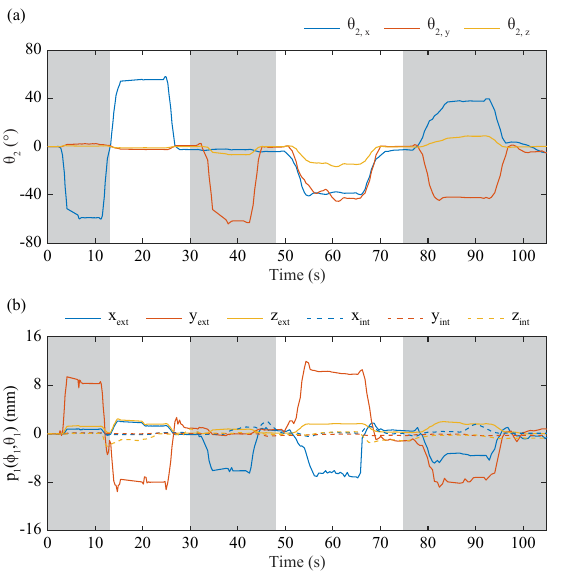}
    \caption{Example of the temporally recorded data for an instance where $\theta_{2}=60^{o}$, showing (a) the three components of angular rotation of the distal segment varying for each pose being performed in sequence, and (b) the effect on the tip position of the proximal segment for both externally and internally routed cabling. Each shaded region indicates the poses displayed in Fig. \ref{stiff_vs_flex}(a)-(e) in sequential order. }\vspace{-0.5cm}
    \label{timeseq}
\end{figure}

\subsection{Stability Analysis}

\begin{figure}[t!]
    \centering
     \includegraphics[width=\linewidth]{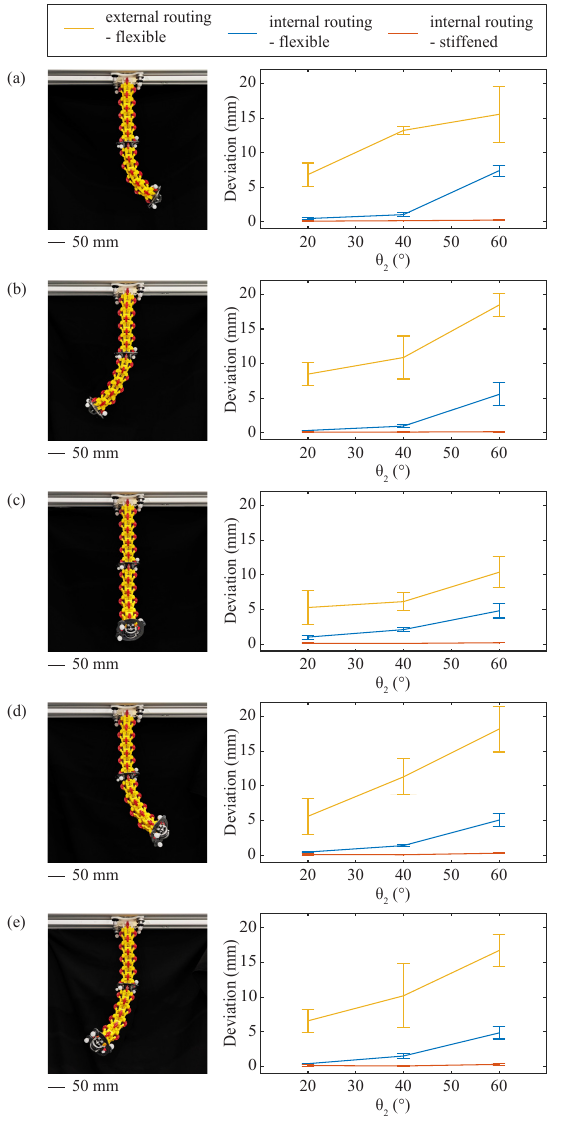}
    \caption{Euclidean deviation observed at the tip of the proximal segment when actuating the distal segment for five different bending plane angles relative to the x-axis: (a) $\phi_2 = 90^{o}$, (b) $\phi_2 = -90^{o}$, (c) $\phi_2 = 0^{o}$, (d) $\phi_2 = 45^{o}$ and (e) $\phi_2 = -45^{o}$. Each bending plane angle was tested with a curvature angle of $\theta_2 = 20^{o}$, $\theta_2 = 40^{o}$ and $\theta_2 = 60^{o}$. } 
    \label{stiff_vs_flex}\vspace{-0.5cm}
\end{figure}

One of the key motivations behind the alternate routing of the driving tendons in contrast to typical continuum tendon driven manipulator design \cite{Webster_Jones_2010} stems from the need for improved stability, intended here as minimizing displacement of the stationary proximal segment while manoeuvring the distal one. To test this feature, 3 different configurations (external routing, internal routing when flexible and internal routing when stiffened) were analysed across 5 different poses, as shown in Fig. \ref{stiff_vs_flex}: these encompass both in-plane bending and out-of-plane bending, covering $\phi_2 = 0^{o}$, $\phi_2 = \pm 90^{o}$ and $\phi_2 = \pm 45^{o}$. The external cable routing is used as a baseline against which the proposed internal routing is assessed. The additional set of tests performed with both internal tendon routing and when the manipulator is in a stiffened state were undertaken to analyse potential stability improvements when combining these features. The deviation in Euclidean distance of the proximal segment tip was recorded for each pose across varying distal segment curvature angles of $\theta_2 = 20^{o}$, $\theta_2 = 40^{o}$ and $\theta_2 = 60^{o}$, performing 3 iterations of each test to monitor variability. 

A temporal sequence of the manipulator response under motion of the distal segment is portrayed in Fig. \ref{timeseq}. This highlights the transition from one configuration of the end-effector to the other, Fig. \ref{timeseq}(a), as shown in the left-hand-side insets of Fig. \ref{stiff_vs_flex}(a) to (e). It also shows the corresponding reaction of the proximal segment in terms of displacement from its original posture, Fig. \ref{timeseq}(b), either with externally (solid line) or internally (dashed line) routed cables. Repetition of these tests provides a statistical representation of the structural response of the system across the configurations tested, Fig. \ref{stiff_vs_flex}. When operating the distal segment with externally routed tendons and in a flexible state, the position of the proximal segment significantly deviates from the initial configuration. Across the set of tested postures, the proximal tip position displays a mean Euclidean deflection of 6.55mm, 10.34mm and 15.86mm for $\theta_2 = 20^{o}$, $\theta_2 = 40^{o}$ and $\theta_2 = 60^{o}$ respectively, relative to the initial state of $q_i = [0, 0]^T$. A maximum deflection of 8.47mm, 13.22mm and 18.47mm was observed for $\theta_2 = 20^{o}$, $\theta_2 = 40^{o}$ and $\theta_2 = 60^{o}$ respectively. When adopting the internally routed method proposed in this design, these values were reduced significantly to 0.53mm, 1.39mm and 5.52mm, a relative decrease of 93.74\%, 89.49\% and 70.11\%. Finally, by incorporating the ability to stiffen through application of tension along the manipulator backbone, these values can be decreased even further to 0.09mm, 0.09mm and 0.23mm, almost zero deviation. These observations clearly indicate that by combining internally routed cables with the ability to stiffen segments independently, the configuration stability of the proximal segment can be greatly improved when operating the distal segment. This forms an attractive characteristic when operating multi-segment continuum manipulators, as the user can be confident that tasks can be undertaken with the end-effector without compromising the desired configuration, a trait inherent in rigid-link manipulator designs.
\begin{figure}[t!]
    \centering
    \includegraphics[width=\linewidth]{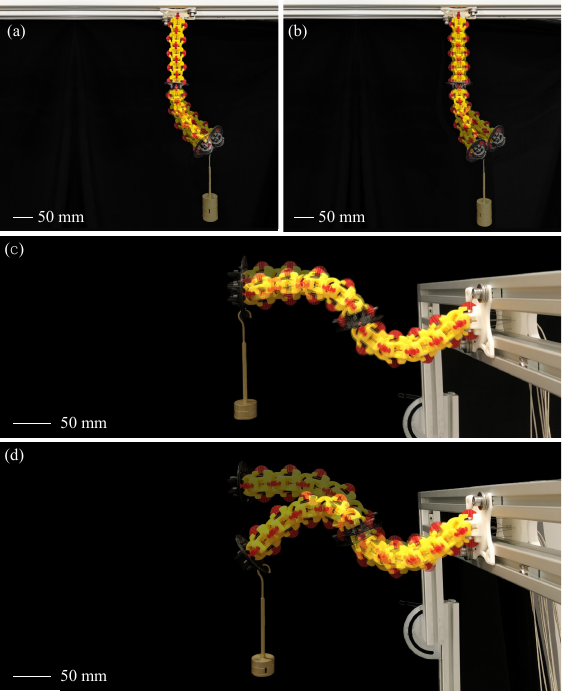}
    \caption{Example of a load bearing test, showing deflection in the vertical configuration (with 500g mass) (a)(b) and the horizontal direction (with 200g mass) (c)(d). Displayed are the stiffened state (a)(c) and the flexible state (b)(d). Evidently, the stiffened state retains its shape more effectively.}\vspace{-0.5cm}
    \label{loading_example}
\end{figure}

Variability of the error, as demonstrated by the spread of the error bars in Fig. \ref{stiff_vs_flex}, also demonstrates higher repeatability of the experiments when stiffened. With externally routed cables, the variation across the 3 tests for each angle is more pronounced with mean ranges of 4.03mm, 4.5mm and 5.41mm for distal curvature angles of $\theta_2 = 20^{o}$, $\theta_2 = 40^{o}$ and $\theta_2 = 60^{o}$ respectively; this is also found to increase with increasing curvature. In contrast, with internally routed cables and stiffening, this variability is drastically reduced to the point of being hardly measurable. This feature highlights the advantages of adopting internally routed tendons, improving the repeatability element of the manipulator control and maintaining reliability across different postures.


\subsection{Load Bearing Capacity}

As previously discussed, soft and continuum manipulators generally lack the ability to bear significant load at the end-effector of the manipulator: this has been shown to prevent the realization of large scale soft manipulators as large disturbances at the tip can detrimentally affect all segments. 

Assessing the load-bearing robustness of a manipulator in response to controllable stiffness is critical for the design and operation of industrial-scale soft robots. With respect to this, the manipulator was tested in a stiffened state by subsequently applying increasing load at the end-effector across 2 poses with $\phi = 0^{o}$ and $\phi = 45^{o}$, and a constant curvature angle of $\theta = 45^{o}$, thus spanning both vertical and horizontal configurations (i.e. when the manipulator backbone is both aligned with gravity and perpendicular). In these experiments, the manipulator was loaded until a ``point of failure" was reached; this effectively constitutes the point where significant deflection is witnessed at the tip. An example of applying the load is shown in Fig. \ref{loading_example}, portraying the (a) stiffened case and (b) flexible case with the latter exhibiting much higher deflection and instability in the proximal segment.
\begin{figure}[t!]
    \centering
    \includegraphics[width=\linewidth]{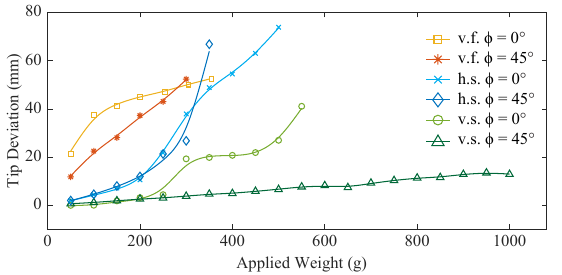}
    \caption{Euclidean tip deviation relative to the load applied, where v. and h. refer to vertical and horizontal configurations respectively; f. and s. refer to flexible and stiff states.}
    \label{load_capability}\vspace{-0.5cm}
\end{figure}

The results from this analysis are displayed in Fig. \ref{load_capability}, confirming that stiffening the manipulator produces the desired effect, whereby a greater load can be held before substantial deflection. For the vertical postures, a marked difference was observed when bending is occurring in-plane (i.e. along the x or y axes) or out-of-plane (i.e. $\phi$ is some value other than a multiple of $\pi/2$). Remarkably, when the distal segment is configured with $q=[\pi/4, \pi/4]^T$ the manipulator is capable of supporting a 1kg load whilst only deviating $\approx 13$mm; it is worth noting that this deviation was incurred successively, with each $50$g increase in load causing a slight increase in deviation. When bending in-plane, this value is much lower, with a significant inflection at $300$g. This outstanding load-bearing capability during out-of-plane bending is motivated by the discretized nature of the manipulator. When jamming is activated out-of-plane, both the top and side of the bead rails are forced into contact, in contrast to only the top when jamming in-plane. It is postulated that this additional shearing contact vastly increases the friction, which in-turn creates a configuration-dependent passive structural support. This is supported by the fact the flexible case with $\phi = 45^{o}$ also displays lower tip deviation in comparison to the same loading for the case with $\phi = 0$. On the contrary, a minor drawback of the discretised design is the reduction in contact area between beads when jamming, thus a higher tensile force must be applied to the tendons to achieve similar holding torques to those noted in \cite{Mukaide2020Radial-LayerConfiguration}.


When positioning the manipulator horizontally, with gravity now acting perpendicular to the manipulator backbone, both in-plane and out-of-plane configurations show similar behaviour and a load of 200g was the threshold in these cases. Despite unavoidable ``sagging" when horizontal, the general workspace of the manipulator was largely unaffected with the maximum reach only reducing by $20\%$ compared to the vertical configuration.
The inference from these results is that the jamming ability inherent in our design can facilitate larger scale continuum designs by improving the load bearing ability from base to tip.

\section{CONCLUSIONS}

This work has presented the modular design and experimental testing of a tendon driven, bead-jamming manipulator with discretised spatial motion capabilities and distributed stiffening. The design features facilitate improved stability of the proximal segment whilst controlling the distal segment through an internally routed tendon mechanism which eliminates unwanted deviations from the desired posture. Also, the load bearing stability of the manipulator was improved by compressing the modular beads together, leveraging frictional effects through incorporation of flexible TPU hinges. For the case where bending occurs out-of-plane ($\phi = 45^{o}$), the manipulator was able to hold up to 1kg whilst only exhibiting $\approx 13$mm tip deviation, a remarkable improvement over both the flexible and stiffened in-plane bending cases ($\phi=0$). Mobility was assessed in both vertical and horizontal configurations, with minimal variation in the achievable workspace. 

Overall, the findings in this work provide evidence that incorporating these features can potentially assist in the development of large, yet compliant manipulators. In future work, the prototype will be expanded and substantially increased in length, aiming to manufacture functional designs in the order of metres. By doing so, the aim is to develop solutions for an area that soft and continuum robotics has lacked until now: large scale, high payload capable manipulation.

\addtolength{\textheight}{-12cm}   








\bibliographystyle{ieeetr}
\bibliography{biblio}

\end{document}